\documentclass[10pt,twocolumn,letterpaper]{article}

\usepackage[pagenumbers]{cvpr} 

\usepackage[dvipsnames]{xcolor}

\usepackage{svg}
\usepackage{graphicx}
\usepackage{amsmath}
\usepackage{amssymb}
\usepackage{booktabs}
\usepackage{bbm}
\usepackage{xcolor}
\usepackage{color, soul}
\usepackage{makecell}
\usepackage{multirow}
\usepackage{booktabs}
\usepackage{caption}

\definecolor{cvprblue}{rgb}{0.21,0.49,0.74}
\usepackage[pagebackref,breaklinks,colorlinks,citecolor=cvprblue]{hyperref}
\usepackage[accsupp]{axessibility}

\def\paperID{11735} 
\def\confName{CVPR}
\def\confYear{2024}

\title{CuVLER: Enhanced Unsupervised Object Discoveries through Exhaustive Self-Supervised Transformers}

\author{Shahaf Arica\qquad Or Rubin\qquad Sapir Gershov\qquad Shlomi Laufer\\
Technion - Israel Institute of Technology\\
{\tt\small shahaftech@gmail.com}
}

\begin{document}
\maketitle
\begin{abstract}
In this paper, we introduce VoteCut, an innovative method for unsupervised object discovery that leverages feature representations from multiple self-supervised models. VoteCut employs normalized-cut based graph partitioning, clustering and a pixel voting approach. Additionally, We present CuVLER (Cut-Vote-and-LEaRn), a zero-shot model, trained using pseudo-labels, generated by VoteCut, and a novel soft target loss to refine segmentation accuracy.  
Through rigorous evaluations across multiple datasets and several unsupervised setups, our methods demonstrate significant improvements in comparison to previous state-of-the-art models. Our ablation studies further highlight the contributions of each component, revealing the robustness and efficacy of our approach. Collectively, VoteCut and CuVLER pave the way for future advancements in image segmentation.
The project code is available on GitHub at \url{https://github.com/shahaf-arica/CuVLER}
\end{abstract}
    
\section{Introduction}
\label{sec:intro}

Object localization remains a cornerstone in computer vision, empowering AI systems with abilities ranging from perception and inference to strategic planning and interactions centered around objects. The conventional training paradigm for such models often requires specialized annotations—be it object bounding boxes, masks, or localized keypoints. Unfortunately, acquiring these manual annotations is time-consuming and resource-heavy \cite{zhao2019object}. Consequently, there is a burgeoning interest in automated object detection and segmentation, particularly in unsupervised settings, circumventing the exhaustive annotation process \cite{zou2023object}.

Wang \etal \cite{wang2023cut} introduced CutLER (Cut-and-LEaRn), a new approach for training unsupervised object detection and segmentation models. CutLER employs a two-step process. First, it generates pseudo-labels using the MaskCut method. This novel approach leverages a \textit {single} self-supervised model to create a \textit {fixed} number of pseudo-labels per image. In the second phase, these pseudo-labels train a segmentation model, resulting in the CutLER model. 
Additionally, CutLER exhibits potential as a base model for supervised detection, demonstrating efficacy in few-shot benchmarks.

Our research builds on this pioneering work, progressing through three critical stages: (1) We employ our innovative method, referred to as 'VoteCut’, to harness feature representations from \textit{multiple} Vision Transformers (ViTs) \cite{dosovitskiy2020image} trained in a self-supervised manner \cite{caron2021emerging, oquab2023dinov2} to generate pseudo-labels with corresponding confidence scores, aided by the eigenvectors of Normalized Cuts (NCut) \cite{shi2000normalized}; (2) The generated pseudo-labels are then used to train a robust object detector, which we refer to as "CuVLER"; (3) The output from this detector aids mask refinement in a separate domain, involving the generation of pseudo-labels from detector predictions in this new domain, followed by filtering. These refined pseudo-labels undergo a subsequent retraining phase, encapsulating our self-training approach.

\noindent Our methodology, the proposed CuVLER model, distinguishes itself from Wang \etal's work in its capability to generate superior object masks and detections in a given domain, without the need of several "in domain" self-training stages. We achieve this by integrating insights from \textit{multiple} ViT models, enhancing the cluster separation process. We also pioneer a self-training strategy that operates within the original and the target domain, unlike the approach in CutLER, which restricts self-training to the original domain. This innovation allows our model to enter self-training stages beyond its initial domain, achieving significant enhancements after just one epoch, showcasing its efficiency and flexibility.

This paper underscores the following contributions in the domain of unsupervised object detection and segmentation:

\begin{enumerate}
    \item \textbf{In-Domain Mask and Detection Discovery - VoteCut}: At the heart of our work is a novel method for identifying high-quality masks and detections within a specific domain. We considerably elevate object localization and segmentation's efficacy by harnessing \textit{multiple} self-supervised models, paving the way for future explorations. Notably, unlike MaskCut \cite{wang2023cut}, these masks are equipped with a confidence score, enhancing their reliability and utility.
    \item \textbf{Instance-Level Loss Function with Soft-Targets}: We present a unique loss function that operates at the instance level and integrates soft-targets. This innovation facilitates a more granular training regimen, boosting object segmentation and detection.
    \item \textbf{Cross-Domain Learning via Self-Training}: Highlighting our method's adaptability, we delineate how our distinctive loss function can be harnessed both within and outside its original domain in a self-training context. Such versatility underscores our model's potential to be adapted across diverse applications, enriching the unsupervised object detection and segmentation landscape.
\end{enumerate}
\section{Related work}
\label{sec:relatedwork}

\noindent\textbf{Self-supervised feature learning.}\hspace{0.5em} Self-supervised learning aims to generate rich data representations without reliance on human annotations, typically achieved through pretext tasks. A noteworthy advancement in this domain has been the training of Vision Transformers (ViTs) \cite{dosovitskiy2020image} in a self-supervised manner, which yields high-quality features. Broadly, pretext tasks fall into two categories: Augmentation-based and Reconstruction-based. Augmentation-based methods posit that varying augmentations of a single sample should produce semantically similar outcomes compared to disparate dataset samples. This often takes the shape of distinguishing augmentations from unrelated samples via constructive \cite{chen2020simple, he2020momentum, misra2020self, wu2018unsupervised}, similarity \cite{chen2021exploring, grill2020bootstrap}, clustering \cite{asano2019self, caron2020unsupervised, xie2016unsupervised}, or category-based \cite{caron2021emerging, oquab2023dinov2} feature learning. Reconstruction-based methods, on the other hand, emphasize reconstructing hidden patches or pixels, aiming to discern object structures within the image \cite{bao2021beit, doersch2015unsupervised, he2022masked}.

\noindent\textbf{Unsupervised instance segmentation.}\hspace{0.5em}
Attaining unsupervised instance segmentation, as demonstrated by FreeSOLO \cite{wang2022freesolo}, involves the extraction of preliminary coarse object masks, followed by mask refinement through a self-training procedure. While FreeSOLO can generate multiple coarse masks per image, their quality occasionally falls short. Similar to our approach, other techniques harnessed DINO \cite{caron2021emerging}-extracted features to pursue instance segmentation. These endeavors are motivated by the observation that DINO features encapsulate meaningful interconnections between patches within each image. Models like LOST\cite{simeoni2021localizing} and TokenCut\cite{wang2022tokencut} leverage self-supervised ViT features for segment discovery through a graph constructed from patch key features' similarity matrix connections. However, their emphasis often remains restricted to the image's primary salient object. Conversely, MaskDistill \cite{van2022discovering} extracts class-agnostic initial masks from a self-supervised DINO's affinity graph, yet its single-mask approach during distillation significantly restricts multi-object detection. CutLER \cite{wang2023cut} has indeed carved a significant mark in object detection and segmentation; however, our method's novelty stems from leveraging multiple models and achieving exceptional results without the need for extensive "in domain" self-training stages, making it a promising advancement in the field of unsupervised object detection and segmentation.
\section{Method}
\label{sec:method}
This study introduces an innovative approach for unsupervised object detection and segmentation using the "cut-vote-and-learn" pipeline. This technique capitalizes on the findings from recent research \cite{simeoni2021localizing,wang2022tokencut,wang2023cut} highlighting the effectiveness of self-supervised representations for object discovery. 
Our pipeline, illustrated in \Cref{fig:VoteCut clustering}, presents a straightforward technique capable of detecting multiple objects, resulting in substantial improvements in segmentation and detection performance within the target domain. Specifically, we first introduce VoteCut, which generates multiple binary masks per image using self-supervised features from DINO \cite{caron2021emerging} in the ImageNet domain \cite{deng2009imagenet}.
We then leverage a loss function with soft targets to enable self-training with these masks. From this point forward, VoteCut with an additional self-training process using our novel loss function will be called "CuVLER". Additionally, we also enhanced CuVLER performance through self-training across different domains.

\subsection{Normalized Cuts}
Normalized Cuts (NCuts) \cite{shi2000normalized} is a popular algorithm for image segmentation and clustering. According to this technique, we represent each patch as a node, thus constructing an undirected, fully connected graph. Each pair of nodes within this graph is connected by a weighted edge that measures their similarity. 
The NCut algorithm attempts to minimize the cost of partitioning the aforementioned graph into sub-graphs by solving a generalized eigenvalue problem:
\begin{equation} \label{eq:eigenvalue}
(D - W)x = \lambda Dx 
\end{equation}

Where W and D denote the adjacency matrix and the degree matrix of the weighted graph, respectively \cite{bronstein2021geometric}. The solution denoted as $x$ in \cref{eq:eigenvalue}, corresponds to the eigenvector associated with the second smallest eigenvalue $\lambda$.
Traditionally, the formulation fixes the number of clusters in the graph, usually a bipartition; however, we preferred a more relaxed approach that utilizes the K-means algorithm \cite{dhillon2004kernel, tepper2011automatically}.

\subsection{VoteCut for object discoveries}
\begin{figure*}
     \centering
     \begin{subfigure}[b]{1.0\textwidth}
         \centering
         \includegraphics[width=\textwidth]{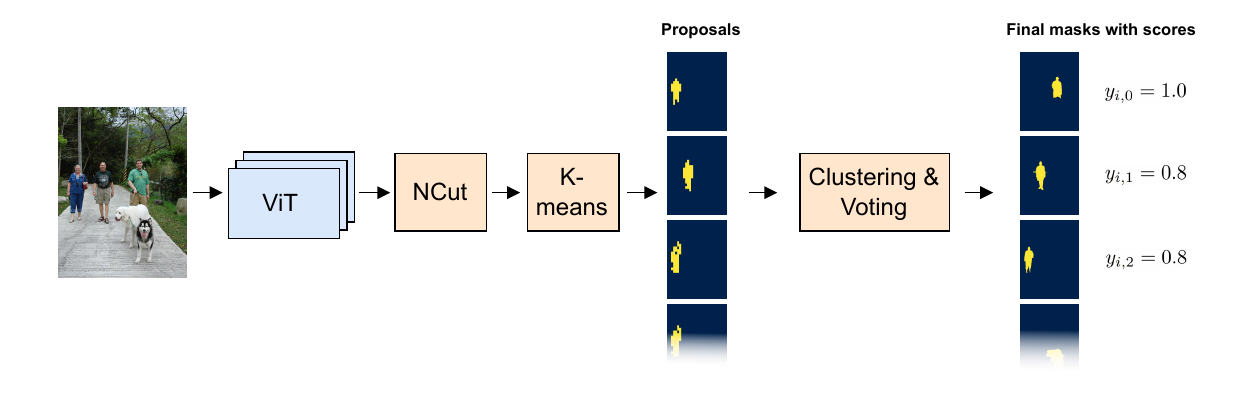}
         \caption{}
         \label{fig:VoteCut pipline}
     \end{subfigure}
     \begin{subfigure}[b]{1.0\textwidth}
         \centering
         \includegraphics[width=\textwidth]{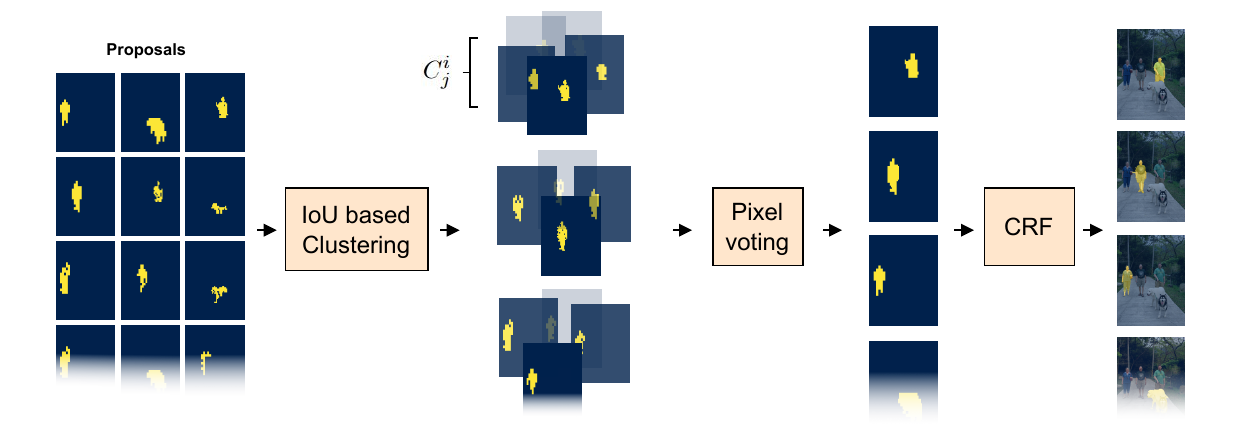}
         \caption{}
         \label{fig: frames optical flow}
     \end{subfigure}
        \caption{\textbf{(a)} An illustrated overview of the VoteCut workflow. A set of models initially makes inferences on the input image, producing feature representations for individual patches. Subsequently, Normalized Cuts (NCut) are performed following the methodology in \cite{wang2022tokencut}, yielding the second smallest eigenvectors from each model. Multiple segment proposals are generated by applying 1D K-means clustering to these eigenvectors with varying K values. The final stage of VoteCut involves clustering these proposals and extracting definitive masks from each cluster via voting. Each definitive mask is also associated with a score. \textbf{(b)} The "Clustering \& Voting" stage of VoteCut is detailed. First, segments are clustered using an Intersection over Union (IoU) threshold, determining segment membership within clusters. A voting mechanism is employed within each cluster to decide whether each patch should be included in the segment. Lastly, a Conditional Random Field (CRF) \cite{krahenbuhl2011efficient} is applied to refine the mask at a finer level. The cluster size determines the score assigned to each mask, as elucidated in \cref{eq:score}.}
        \label{fig:VoteCut clustering}
\end{figure*}

This work presents a novel technique designed to harness the collective power of multiple self-trained models via a voting mechanism, leading to precise object segmentation (see \Cref{fig: frames optical flow}). 
We capture diverse image content perspectives using an ensemble of models trained on different augmented image sets. With their varying transformer patch sizes, these models can focus on distinct image attributes, thus maximizing object detection precision.  
Our proposed method maximizes the collective intelligence of the aforementioned models by conducting a voting procedure on each image segment. It prioritizes the most widely agreed-upon masks while diminishing the influence of masks with fewer votes.

In line with the methodology presented in TokenCut by Wang et al. \cite{wang2022tokencut}, we adopt a procedure involving the extraction of the second smallest eigenvector from each model, as determined by the Normalized Cut (NCut) algorithm \cite{shi2000normalized}, for each input image. 

In this study, we employ this approach on multiple DINO and DINOv2 \cite{oquab2023dinov2} models, which exhibit different patch sizes, to obtain feature representations for individual patches. Subsequently, these representations are used to construct the affinity matrix employed in the NCut algorithm. To calculate elements in the affinity matrix $W$, we use the cosine similarity between the patch’s features. For DINO models, we follow Wang \etal \cite{wang2023cut} and use the 'key' features extracted from the endmost attention layer. When using a DINOv2 model, the features used instead are the output features of the endmost attention layer. The calculation is detailed in \Cref{eq:similarity_weights}, where $K_i$ is the ‘key’ feature of the $i$‘th patch and $f_i$ is the output feature of the $i$’th patch. We follow Wang \etal \cite{wang2023cut} and apply a threshold operation to the elements of matrix $W$. Specifically, we set any $W_{ij} \geq \tau^{\text{ncut}}$ to 1, and otherwise to $1e^{-5}$.
\begin{equation}
\label{eq:similarity_weights}
W_{ij}=\begin{cases}
\frac{K_{i}K_{j}}{\Vert K_{i}\Vert_{2}\Vert K_{j}\Vert_{2}} & \text{DINO model is used}\\
\frac{f_{i}f_{j}}{\Vert f_{i}\Vert_{2}\Vert f_{j}\Vert_{2}} & \text{DINOv2 model is used}
\end{cases}
\end{equation}

Subsequently, we generate mask proposals by applying 1D K-means clustering \cite{arthur2007k} to the eigenvectors using every k value, ranging from 2 to $k_{\text{max}}$. This process resulted in the creation of $n$ instance mask proposals per image, achieved by applying connected-components analysis to each segment produced by the K-means clustering. 

Following this, we utilized an intersection over union (IoU)-based strategy to group masks into clusters. Given $n$ proposal masks for a specific image, our clustering procedure begins with a greedy selection process. For each iteration, we identify the instance mask with the highest number of overlaps with other masks, surpassing an IoU threshold of $\tau^c=0.6$. This mask is designated as the cluster pivot. The masks that share substantial IoU overlap with the pivot mask are considered part of the same cluster. Once a new cluster is formed, we systematically remove all associated masks and repeat this clustering procedure iteratively. This recursive process persists until all instance masks have been effectively associated with their respective clusters.

Formally, we denote the collection of clusters for the $i$-th image in the image set as \( \mathcal{C}^i = \{C_1^i, C_2^i, \ldots, C_m^i\} \). In the context of each cluster \( C_j^i \), we designate the mask members as \(\{M_1^i, M_2^i, \ldots, M_p^i\}\). The resulting final mask for the cluster is determined as follows:

\begin{equation}
\text{Final Mask} = \begin{cases}
1, & \text{if } \frac{1}{p}\sum_{k=1}^{p} M_k^i > \tau^m \\
0, & \text{otherwise}
\end{cases}
\label{eq: Final_Mask_Decision}
\end{equation}
Here, \(\tau^m\) represents a threshold, and the final mask is set to 1 if the average of all masks for a given pixel within the cluster exceeds this threshold. \(\tau^m\) sets the required consensus among the majority of masks to result a value of \(1\) in the final mask. We leverage a Conditional Random Field (CRF) \cite{krahenbuhl2011efficient} to perform post-processing on the final masks, facilitating the computation of their associated bounding boxes.

Once the clustering procedure is over, we compute the mask proposal score. This score, ranging from 0 to 1, is provided to each cluster \(j\) in each image \(i\) in the image set and is denoted by \(y_{i,j}\). It corresponds to the cluster yielding the highest consensus mask value among all mask proposals of the same cluster size:

\begin{equation}
     y_{i,j} = \frac{\left|C_j^i\right|}{\max\left(\left|C_1^i\right|, \left|C_2^i\right|, \ldots, \left|C_{m}^i\right|\right)}
  \label{eq:score}
\end{equation}

As demonstrated in the following section, we showcase the applicability of this score in training a model using our innovative loss function applied at the instance level. This approach enables us to utilize all suggested masks without concern for inaccuracies, as those with lower scores will have a minor impact on the model's performance.

Based on this procedure, many clusters of VoteCut receive a score close to zero and, as such, have minimal contribution to the loss function. To minimize the calculation time, in cases where there are more than 10 VoteCut clusters, we remove the masks with the lowest scores.

\subsection{Soft loss function}
Given the aforementioned $y_{i,j}$, the score for the $j$-th pseudo-labeled instance in the $i$-th image, the corresponding bounding box loss is formulated as follows:

\begin{equation}
     L_{box}=\sum_{i \in I}\sum_{j \in G_i} y_{i,j}L_{box, orig}^j
  \label{eq:important}
\end{equation}

Here, $I$ denotes the image set, $G_i$ denotes the set of instances (i.e., masks) that are associated with the $i$-th image, and $L_{box, orig}^j$ is the original loss for the $j$-th box of the instance using input image $x_i$. Similarly, the mask loss is defined as:

\begin{equation}
     L_{mask}=\sum_{i \in I}\sum_{j \in G_i} y_{i,j}L_{mask, orig}^j
  \label{eq:important}
\end{equation}

For foreground score, a soft binary cross-entropy is employed:

\begin{equation}
     L_{cls}=\sum_{i \in I}\sum_{j \in G_i} y_{i,j}\log(\sigma_f(x_i))+(1-y_{i,j})\log(\sigma_b(x_i))
  \label{eq:important}
\end{equation}

Where $\sigma_f(x_i)$ represents the softmax output for foreground and $\sigma_b(x_i)$ for background.

Lastly, inspired by Wang et al. DropLoss \cite{wang2023cut}, we define $r_{i,j}$ as the predicted region with maximum overlap of $\tau^{\text{IoU}}$ against 'ground truth' instances. This leads to the comprehensive loss function:

\begin{equation}
     L=\sum_{i \in I}\sum_{j \in G_i}\mathbbm{1}(\text{IoU}_{r_{i,j}}^{max} > \tau^\text{IoU})(L_{cls}+L_{box}+L_{mask})
  \label{eq:important}
\end{equation}

Where $\text{IoU}_{r_i,j}^{max}$ signifies the highest IoU with all pseudo-labeled instances. This loss avoids penalizing the model for missing 'ground-truth' objects, fostering exploration of diverse image regions. Similarly to Wang et al.'s work \cite{wang2023cut}, we applied a low threshold of $\tau^\text{IoU}=0.01$.

\subsection{CuVLER}
\label{subsec:CuVLER}

Following the initial training stage, we conduct class-agnostic detection following the methodology outlined in \cite{simeoni2021localizing}. This involves training a detector in a class-agnostic manner (CAD), utilizing the masks and the scores generated by VoteCut. It is essential to notice that CuVLER is trained solely on the ImageNet validation dataset. Thus, different datasets, such as the COCO dataset, can be considered candidates for zero-shot performance evaluation.

\subsubsection*{Self-training on a different domain}
To generate pseudo-labels for a new dataset, we produce them using the CuVLER model inference. Then, we filtered out instances with confidence scores lower than 0.2. This newly curated dataset is utilized for further training, incorporating the updated confidence scores within our soft loss framework, resulting in the refinement of mask predictions.

\subsection{Implementation details}
\label{subsec: implemntation-details}
\textbf{Pre-processing stage}
When using DINO models, we resized the images to 480x480 pixels, and when using DINOv2 models, we resized the images to 476x476 pixels. 

\textbf{VoteCut}
The complete list of the utilized models appears in the supplementary materials (see  \cref{subsec:Utilized-models}). Unless specified otherwise, we utilized all aforementioned models and set $\tau^m=0.2$ and $k_{max}=3$. Further ablations related to these hyperparameters can be found in \cref{sec:ablations}. We set $\tau^{cut}=0.15$, as done by Wang \etal \cite{wang2023cut}.

We employ a two-step resizing approach to ensure accurate pixel alignment between proposals. First, we resize all proposals to a fixed size before the IOU clustering phase. Then, after the pixel voting phase, we resize the final mask to match the image shape.

\textbf{Training details}
All experiments were performed using the Detectron2 \cite{wu2019detectron2} platform, using a batch size of 16 and the copy-paste augmentation \cite{Dwibedi_2017_ICCV, ghiasi2021simple}. 
Cascade Mask R-CNN \cite{cai2018cascade} detector is used for CAD.

\section{Experiments}
\label{sec:experiments}

This section delves into our experimental framework and is designed to evaluate our method's performance comprehensively. We divide the experiments into three essential evaluations, each presenting a different side of our methodology. First, we scrutinize the effectiveness of our method 'in domain' - within the domain for which the ViT models have initially trained. This provides us with a performance assessment in a familiar context. Second, we venture 'out of domain' zero-shot evaluation to assess the model's generalization capabilities (i.e., examining its adaptability to new environments). Lastly, we examine the dynamics of 'self-training' within an alternative domain. This evaluates the efficacy of unlabeled images from the domain of interest to enhance object discovery tailored to that domain. These experiments collectively offer a thorough understanding of the strengths and limitations of our approach in diverse settings, providing valuable insights for its practical applications.

The assessment of unsupervised object detectors presents a unique set of challenges. Primarily, these models lack an inherent understanding of semantic classes, rendering them unsuitable for evaluation through class-aware detection metrics. Consequently, we adopt the class-agnostic detection evaluation paradigm in line with prior research \cite{bar2022detreg, simeoni2021localizing, wang2022tokencut, wang2023cut}. Secondly, object detection datasets typically provide annotations for only a subset of the objects present in the images. Similarly to the work of Wang et al. \cite{wang2023cut}, we have incorporated the Average Recall (AR) metric to address this limitation. AR proves valuable in assessing unsupervised detection models as it refrains from penalizing them for detecting novel, unlabeled objects within the dataset.

\subsection{In-domain evaluation}
In this experiment, we conducted an 'in-domain' evaluation of the ImageNet validation set. We chose this specific dataset to align with the domain on which the ViT model was originally trained. Our comparative analysis unfolds through two distinct scenarios: (1) 'no CAD', where we solely generate masks using the features extracted from the pre-trained model, abstaining from training a dedicated detector; (2) 'with CAD', where we train a detector using the masks we create and subsequently deploy it for class-agnostic object detection. Keeping with established conventions, our evaluation employs the widely recognized COCO metrics. However, given that only a fraction (approximately 10 \%) of ImageNet has bounding-box annotations and none have segmentations, we exclusively report the performance of bounding-box metrics.

Our approach introduces a novel scoring mechanism, detailed in \cref{eq:score}, which allows for a more detailed and insightful assessment of our model's performance. To facilitate a fair and equitable comparison, we incorporated DINOv2 models within our methodology. Consequently, we compared the previous state-of-the-art (SOTA) and our best-performing DINOv2 model. For reference, the 'no-CAD' method of the previous SOTA was set to 1.0 since this methodology does not provide a direct score evaluation. 

It's worth noting that this additional comparison does not yield any significant improvement in favor of the previous SOTA; this is shown in \Cref{tab:indomain}, where MaskCut$^\dagger$ utilizes the best-performing DINOv2 model, instead of the DINO model used in the original MaskCut \cite{wang2023cut}.   

As depicted in \Cref{tab:indomain}, in the 'no CAD' scenario, VoteCut (our approach) showcases substantial improvements, with performance enhancements ranging from approximately 60\text{\%} to 100\text{\%} across various metrics. In the 'with CAD' scenario, we observed more modest yet significant improvements, ranging from 4\text{\%} to 13\text{\%} across all metrics. These results underscore our approach's competence and efficiency in 'in-domain' evaluations.

\begin{figure*}
    \centering
    \includegraphics[width=0.61\textwidth]{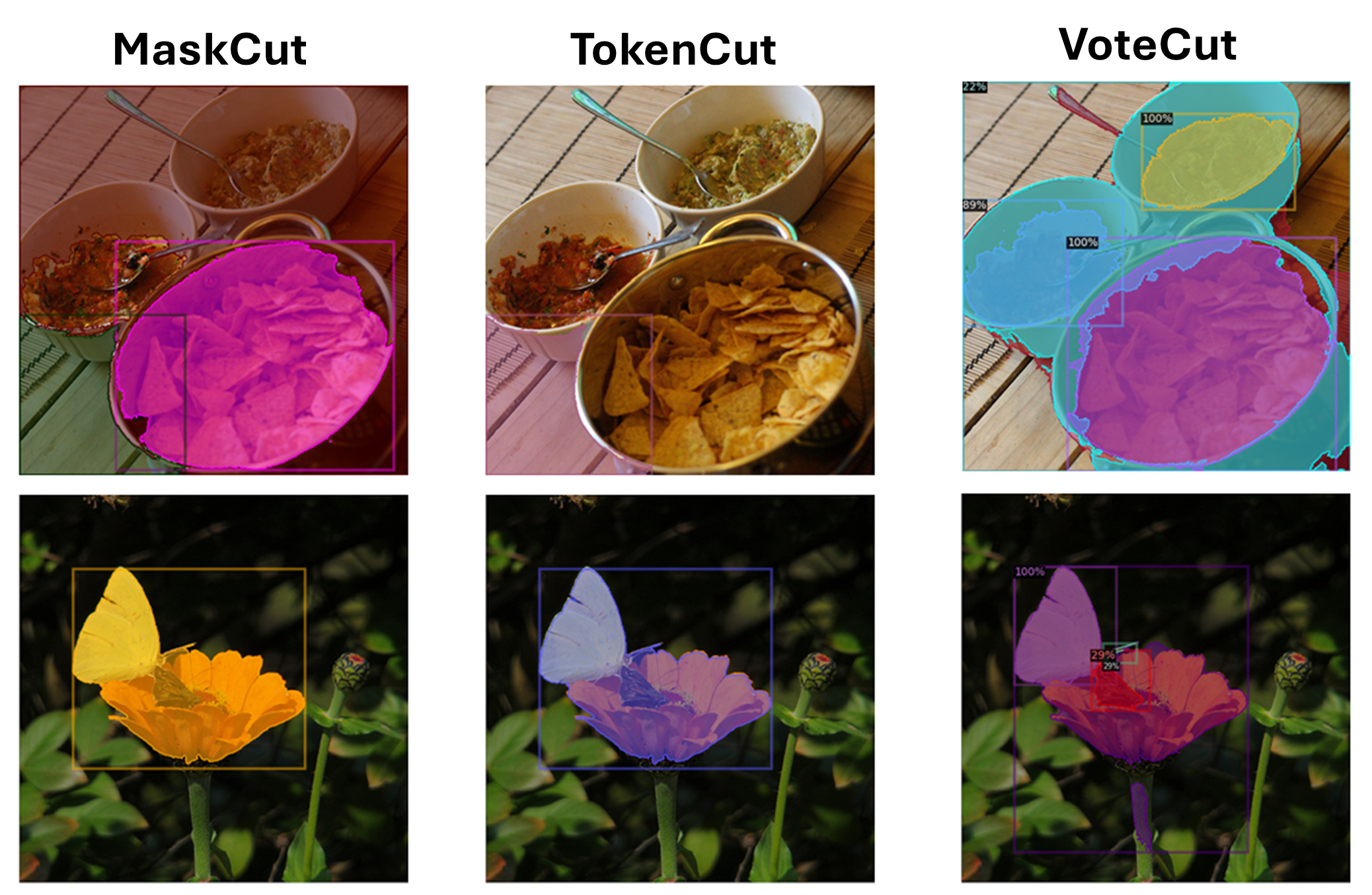}
    \caption{Visual illustration of VoteCut performance vs. SOTA NCut based object-discovery methods on the ImageNet validation set. The VoteCut bounding box score is calculated according to \cref{eq:score}}
    \label{fig:visual_results}
\end{figure*}

Figure \ref{fig:visual_results} illustrates our proposed methodology performance compared to other SOTA models. 

\begin{table}
\centering
\begin{tabular}{p{2.35cm}|c|cccc}
Method&CAD&AP&AP$_{50}$&AP$_{75}$&AR$_{100}$\\
\hline
TokenCut\cite{wang2022tokencut}&$\times$ \quad&14.4&27.0&13.4&26.9\\
MaskCut\cite{wang2023cut}&$\times$ \quad&10.6&20.3&10.0&27.7\\
MaskCut\textsuperscript{$\dagger$}\cite{wang2023cut}&$\times$ \quad&8.3&14.9&7.7&22.8\\
VoteCut (ours)&$\times$ \quad&\textbf{20.9}&\textbf{36.2}&\textbf{20.0}&\textbf{45.0}\\
\hline
CutLER\cite{wang2023cut}&\checkmark \quad&29.2&48.8&29.8&56.7\\
CuVLER (ours)&\checkmark \quad&\textbf{33.2}&\textbf{52.6}&\textbf{33.7}&\textbf{59.0}\\
\hline
\end{tabular}
\caption{In-domain evaluation on the ImageNet validation set. The comparative analysis is divided into two scenarios: ’no CAD’ and ’with CAD’. Keeping with established conventions, the evaluation employs the COCO metrics. We exclusively report the performance of bounding-box metrics. $^\dagger$: utilizing our best-performing DINOv2, instead of the DINO model used in MaskCut \cite{wang2023cut}.}
\label{tab:indomain}
\end{table}

\subsection{Zero-shot evaluation}
In this experiment, we assessed our methodology performance across seven diverse benchmarks, detailed in the Supplementary Materials\nocite{lin2014microsoft, everingham2010pascal, kuznetsova2020open, inoue2018cross}. Following the methodology of the previous SOTA \cite{wang2023cut}, we employ a cascade Mask R-CNN model trained exclusively on ImageNet, a methodology referred to as 'zero-shot' due to its singular domain training and cross-domain evaluation without further adaptation. Our evaluation is based on the COCO metrics, encompassing Average Precision (AP) and AP$_{50}$ scores. Detailed results for all benchmarks are available in the Supplementary Materials.

As depicted in \Cref{tab:zeroshot}, our approach demonstrates significant improvements of up to 20\%. We consistently observe performance enhancements across all benchmarks, except for the Cliparat dataset, where a minor decline of 1\% is noted in the $AP_{50}$ metric. Importantly, our method achieves superior performance over the previous SOTA after a single training epoch, obviating the necessity for extensive self-training stages on the ImageNet dataset.

\begin{table*}
\centering
\begin{tabular}{p{2.4cm}|p{0.6cm}p{0.6cm}|p{0.6cm}p{0.6cm}|p{0.6cm}p{0.6cm}|p{0.6cm}p{0.6cm}|p{0.6cm}p{0.6cm}|p{0.6cm}p{0.6cm}|p{0.6cm}p{0.6cm}|}
\multirow{2}{*}{Method}&\multicolumn{2}{c|}{COCO}&\multicolumn{2}{c|}{COCO20K}&\multicolumn{2}{c|}{VOC}&\multicolumn{2}{c|}{OpenImages}&\multicolumn{2}{c|}{Clipart}&\multicolumn{2}{c|}{Watercolor}&\multicolumn{2}{c|}{Comic}\\ \cline{2-15}
&AP$_{50}$&AP&AP$_{50}$&AP&AP$_{50}$&AP&AP$_{50}$&AP&AP$_{50}$&AP&AP$_{50}$&AP&AP$_{50}$&AP\\ \hline
Prev. SOTA \cite{wang2023cut} &21.9&12.3&22.4&12.5&36.9&20.2&17.3&9.7&21.1&8.7&37.5&15.7&30.4&12.2\\
CuVLER (ours)&23.0&12.6&23.5&12.7&39.4&22.3&19.6&11.6&20.8&9.3&41.3&19.0&32.2&14.6\\
\textit{vs. prev. SOTA}&\textbf{+1.1}&\textbf{+0.3}&\textbf{+1.1}&\textbf{+0.2}&\textbf{+2.5}&\textbf{+2.1}&\textbf{+2.3}&\textbf{+1.9}&-0.3&\textbf{+0.6}&\textbf{+3.8}&\textbf{+3.3}&\textbf{+1.8}&\textbf{+2.4}\\
\hline
\end{tabular}
\caption{SOTA zero-shot unsupervised object detection performance on seven datasets. The reported results are based on the COCO metrics, encompassing both Average Precision (AP) and AP$_{50}$ scores. The presented models are trained in an unsupervised manner solely on ImageNet. Results of \cite{wang2023cut} are produced with official code and checkpoint.}
\label{tab:zeroshot}
\end{table*}

\subsection{Self-training evaluation}
In this experiment, we harness our self-training methodology, which leverages unlabeled images from the domain of interest and subjected it to a rigorous comparison with previous approaches. Aligning with established practices, we train a Cascade Mask-RCNN model on the COCO train2017 dataset and evaluate our results on widely recognized benchmarks for bounding-box and segmentation tasks.

In \Cref{tab:selftraincoco}, we present a comparison of our detector's performance on two prominent benchmarks: COCO \texttt{val2017} and COCO \texttt{20K}, the latter being a subset of 20,000 images from the COCO dataset \cite{lin2014microsoft, wang2022freesolo, simeoni2021localizing, wang2023cut}. Notably, our results showcase improvements in all metrics, with enhancements reaching up to approximately 10\text{\%}.

We have expanded our evaluation to a more challenging benchmark - LVIS \cite{gupta2019lvis}, which encompasses over 1,000 entry-level object categories and naturally exhibits a long-tailed data distribution. We found that the improvement was more modest, with enhancements of up to 7\text{\%~} (see \Cref{tab:selftrainlvis}). This outcome aligns with our expectations, considering the significant shift in category distribution between LVIS and ImageNet. Additional insights and detailed explanations regarding these benchmarks are available in the Supplementary Materials.

\begin{table*}
  \centering
  \begin{tabular}{lccc@{\hspace{0.5pt}}c@{\hspace{0.5pt}}c@{\hspace{0.5pt}}c@{\hspace{0.5pt}}c@{\hspace{0.5pt}}c@{\hspace{0.5pt}}c@{\hspace{0.5pt}}c@{\hspace{0.5pt}}c@{\hspace{0.5pt}}c@{\hspace{0.5pt}}c@{\hspace{0.5pt}}c@{\hspace{0.5pt}}}
    \toprule
    & & & \multicolumn{6}{c}{COCO 20K} & \multicolumn{6}{c}{COCO val2017} \\
    \cmidrule(lr){4-9} \cmidrule(lr){10-15}
    Method & Detector & Init. & AP$_{50}^{\text{box}}$ & AP$_{75}^{\text{box}}$ & AP$^{\text{box}}$ & AP$_{50}^{\text{mask}}$ & AP$_{75}^{\text{mask}}$ & AP$^{\text{mask}}$ & AP$_{50}^{\text{box}}$ & AP$_{75}^{\text{box}}$ & AP$^{\text{box}}$ & AP$_{50}^{\text{mask}}$ & AP$_{75}^{\text{mask}}$ & AP$^{\text{mask}}$ \\
    \midrule
    LOST\cite{simeoni2021localizing} & FRCNN & DINO & - & - & - & 2.4 & 1.0 & 1.1 & - & - & - & - & - & -  \\
    MaskDistill \cite{van2022discovering} & MRCNN & MoCo & - & - & - & 6.8 & 2.1 & 2.9 & - & - & - & - & - & -  \\
    FreeSOLO \cite{wang2022freesolo} & SOLOv2 & DenseCL & 9.7 & 3.2 & 4.1 & 9.7 & 3.4 & 4.3 & 9.6 & 3.1 & 4.2 & 9.4 & 3.3 & 4.3  \\
    CutLER \cite{wang2023cut} & Cascade & DINO & 22.4 & 11.9 & 12.5 & 19.6 & 9.2 & 10.0 & 21.9 &  11.8 & 12.3 & 18.9 & 9.2 & 9.7  \\
    CuVLER\textsuperscript{$\dagger$} (ours) & Cascade & DINO & \textbf{24.1} & \textbf{12.3} & \textbf{13.1} & \textbf{21.6} & \textbf{9.7} & \textbf{10.7} & \textbf{23.5} & \textbf{12.0} & \textbf{12.8} & \textbf{20.4} & \textbf{9.6} & \textbf{10.4} \\ 
    \bottomrule
  \end{tabular}
  \caption{ Unsupervised object detection and instance segmentation on COCO \texttt{20K} and COCO \texttt{val2017}. We report the detection and segmentation metrics and note the detectors (Detector) and backbone initialization (Init.). All models results are obtained with the official code and checkpoint. $^\dagger$: model was further self-trained on the target domain.}
  \label{tab:selftraincoco}
\end{table*}

\begin{table}
\centering
\begin{tabular}{p{1.65cm}|p{0.5cm}p{0.5cm}p{0.7cm}|p{0.6cm}p{0.6cm}p{0.6cm}}
Method&AP$^\text{box}$&AP$_{50}^\text{box}$&AP$_{75}^\text{box}$&AP$^\text{mask}$&AP$_{50}^\text{mask}$&AP$_{75}^\text{mask}$\\
\hline
CutLER\cite{wang2023cut}&4.5&8.4&3.9&3.5&6.7&3.2\\CuVLER\textsuperscript{$\dagger$}&\textbf{4.7}&\textbf{8.9}&\textbf{4.1}&\textbf{3.8}&\textbf{7.2}&\textbf{3.4}\\
\hline
\end{tabular}
\caption{Evaluation on the LVIS benchmark.$^\dagger$: model was further self-trained on the target domain.}
\label{tab:selftrainlvis}
\end{table}
\section{Ablations}
\label{sec:ablations}
\begin{table}
\centering
\begin{tabular}{l|cc}
Methods&AP$_{50}^{mask}$&AP$^{mask}$\\
\hline
MaskCut CAD \cite{wang2023cut} &15.8&7.7\\
\hline
+VoteCut CAD&17.7&9.1\\
+Soft target loss (CuVLER) &19.3&9.8\\
+Self-training stage&20.4&10.4\\
\hline
\end{tabular}
\caption{
Component ablation study of our methodology. We illustrate the impact of each component on the COCO \texttt{val2017} dataset.}
\label{tab:ablationparts}
\end{table}

In \Cref{tab:ablationparts}, we present an ablation study on the COCO \texttt{val2017} dataset \cite{lin2014microsoft}, aiming to illustrate the importance of each component. Utilizing VoteCut with CAD training, as an alternative to the baseline MaskCut \cite{wang2023cut}, after CAD training, results in a 12\% and an 18\% improvement in $AP^{mask}_{50}$ and $AP^{mask}$ respectively. Additionally, utilizing the soft target loss further enhances the results, with a notable 9\% increase in $AP^{mask}_{50}$ and 7\% in $AP^{mask}$. We emphasize that the results achieved up to this point were obtained without reliance on the COCO dataset, thus establishing a zero-shot evaluation setup. Furthermore, incorporating data from the COCO dataset using the self-training stage, as detailed in \cref{subsec:CuVLER}, yields an additional 5\% boost for $AP^{mask}_{50}$ and a 6\% improvement in $AP^{mask}$. Collectively, these components produce a noteworthy 29\% increase in $AP^{mask}_{50}$ and a substantial 35\% enhancement in $AP^{mask}$, highlighting their combined impact on segmentation quality.

\begin{figure}
    \centering
    \includegraphics[width=0.9\columnwidth]{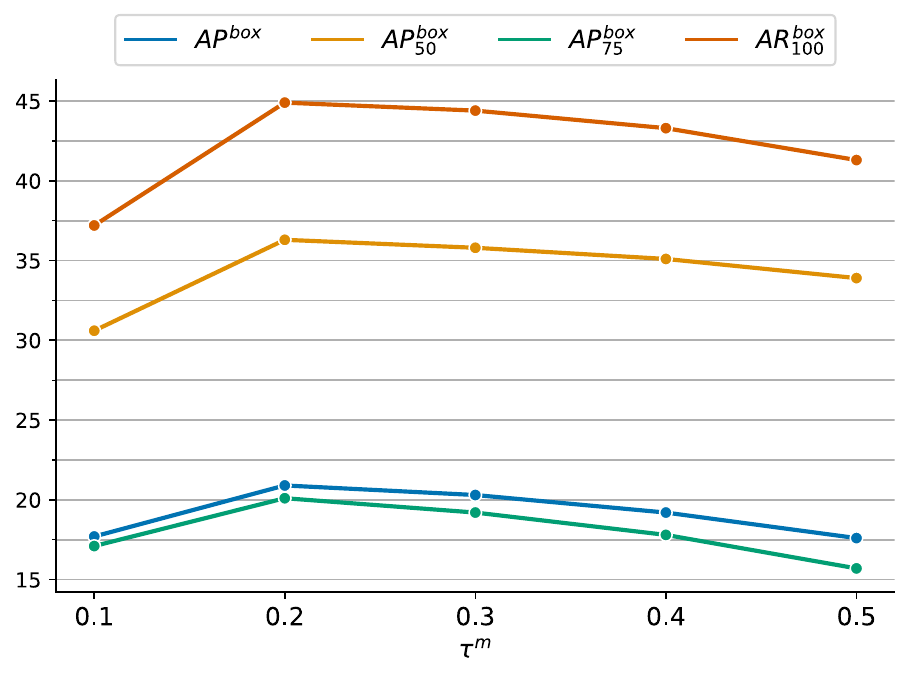}
    \caption{In-domain evaluation of the VoteCut method, without CAD training, with varying $\tau^m$ on the ImageNet validation set.}
    \label{fig:tau_m_ablation}
\end{figure}

We evaluate the VoteCut method (without CAD) in the following ablation studies on the ImageNet validation set. \cite{lin2014microsoft} dataset, in an in-domain setup. These studies aim to investigate the effects of individual hyperparameters while holding the remaining parameters constant. 
\\
$\tau^m$ introduced in \cref{eq: Final_Mask_Decision} acts as an integral threshold within the proposed method. In \Cref{fig:tau_m_ablation}, we demonstrate the impact of varying threshold values. Configuring $\tau^m$ at 0.2 is strictly dominant across all evaluated metrics. 

\begin{figure}
    \centering
    \includegraphics[width=0.9\columnwidth]{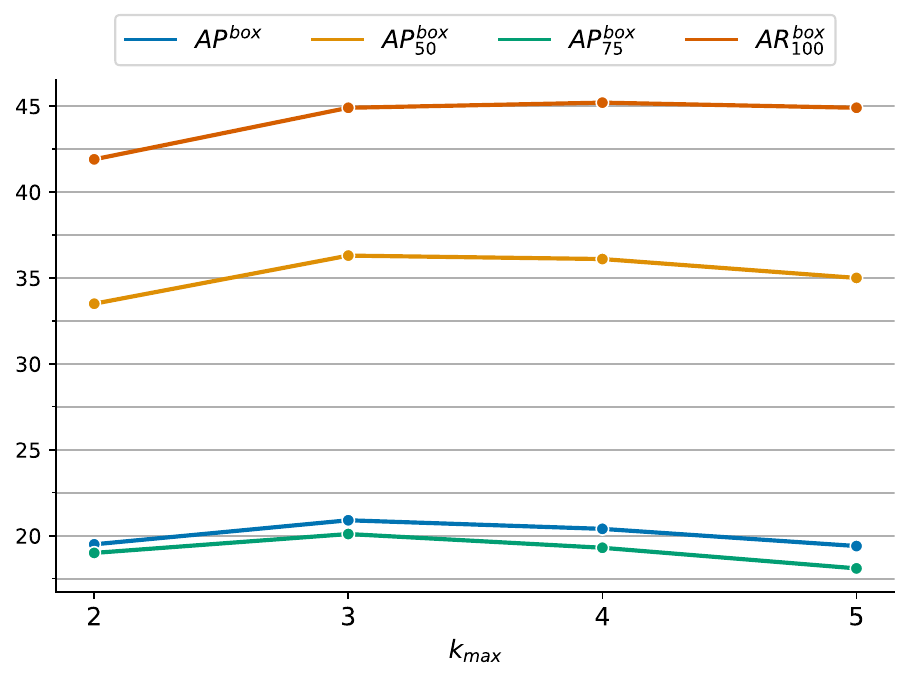}
    \caption{Results of the VoteCut method without CAD training in an in-domain configuration with different $k_{max}$ values on the ImageNet validation set.}
    \label{fig:K_value_ablation}
\end{figure}

$k_{max}$ is an integral part of the VoteCut method, directly impacting the number of generated proposals. In \Cref{fig:K_value_ablation}, we aim to illustrate the influence of varying $k_{max}$ values.  Notably, $k_{max}=3$ excels in AP-based metrics and ranks almost equally well in $AP^{box}$, with under a one percent difference from the top result.

We observe from the provided ablation studies that the proposed method demonstrates robustness to variations in both $\tau^{m}$ and $k_{max}$, as even suboptimal settings for these parameters outperform TokenCut \cite{wang2022tokencut}
by a significant margin.

\begin{figure}
    \centering
    \includegraphics[width=0.95\columnwidth]{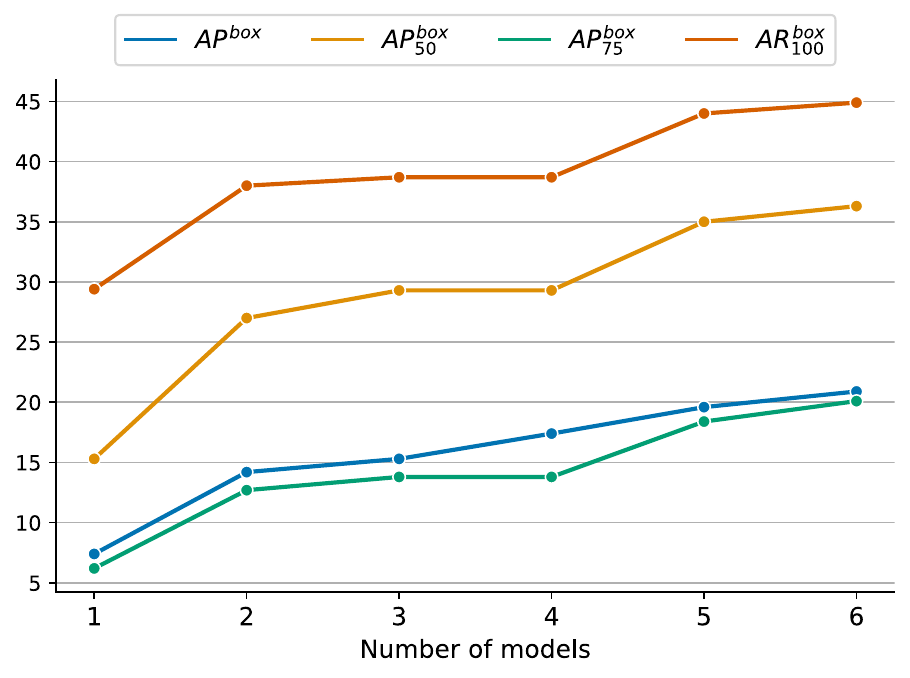}
    \caption{
    Model count ablation test. The results are obtained in an in-domain setup on the ImageNet validation set using the VoteCut method without CAD training.}
    \label{fig:models_num_ablation}
\end{figure}

In \Cref{fig:models_num_ablation}, we can discern the impact of employing an increased number of models. The results correspond to maximum obtained by calculating all possible combinations of models within our set, with each calculation constrained by the number of selected models. A clear trend emerges, indicating that employing a greater number of models leads to improved outcomes. This trend suggests that each model captures distinct information, and the collective strength of the models surpasses the performance of each model in isolation.

\section{Conclusion and Limitation}
\label{sec:conclusion}
We have presented VoteCut, a novel method for unsupervised object discovery capable of generating a dynamic number of mask instances with corresponding confidence scores, outperforming previous counterparts. We also introduced CutLER, a zero-shot model that enhances VoteCut results using a self-training phase with a novel soft loss. Furthermore, we presented an additional enchantment based on self-training that improves cross-domain results. CutLER outperforms the previous SOTA in both zero-shot and unsupervised setups across multiple datasets. 

While our hyperparameter studies highlighted the robustness of our method — with optimal configurations for parameters like \(\tau^m\) and \(k_{max}\) substantially influencing results — we also observed the inherent strength of ensemble techniques. For instance, as the number of models increased, there was a clear trend of performance improvement, signaling the importance of diversified model input.

\textbf{Limitation.} The computational requirements of integrating multiple models within the VoteCut framework may present challenges in resource-constrained settings. However, leveraging VoteCut for training a single segmentation model, as showcased by CuVLER, mitigates this constraint while enhancing segmentation performance. Additionally, utilizing the ImageNet dataset for pseudo-label generation may introduce biases, given its simplified domain and lower object density, potentially reducing false positives. Future research should investigate how source domain characteristics impact pseudo-label quality and inference outcomes.

\newpage
{
    \small
    \bibliographystyle{ieeenat_fullname}
    \bibliography{main}
}
\clearpage
\setcounter{page}{1}
\maketitlesupplementary



\section{Additional details}
\subsection{Utilized models}
\label{subsec:Utilized-models}
\begin{itemize}
    \item DINO ViT-B/8 \cite {caron2021emerging}
    \item DINO ViT-B/16 \cite {caron2021emerging}
    \item DINO ViT-S/8 \cite {caron2021emerging}
    \item DINO ViT-S/16 \cite {caron2021emerging}
    \item DINOv2 ViT-S/14
    \cite
    {oquab2023dinov2}
    \item DINOv2 ViT-B/14
    \cite
    {oquab2023dinov2}
    
\end{itemize}
\subsection{Ablation results}
In \Cref{tab:model_count_ablation}, 
\Cref{tab:k_max-ablation-results} and
\Cref{tab:tau_m_ablation-results} we present the specific results used to create \Cref{fig:tau_m_ablation}, \Cref{fig:K_value_ablation} and \Cref{fig:models_num_ablation}, respectively.
 \begin{table}[!ht]
    \centering
    \begin{tabular}{p{2cm}|p{0.8cm}p{0.8cm}p{0.8cm}p{0.8cm}}
    Model count & $AP^{box}$ & $AP_{50}^{box}$ & $AP_{75}^{box}$ & $AR_{100}^{box}$\tabularnewline
    \hline 
    \hfil 1 & 7.4 & 15.3 & 6.2 & 29.4\tabularnewline
    \hfil 2 & 14.2 & 27 & 12.7 & 38\tabularnewline
    \hfil 3 & 15.3 & 29.3 & 13.8 & 38.7\tabularnewline
    \hfil 4 & 17.4 & 29.3 & 13.8 & 38.7\tabularnewline
    \hfil 5 & 19.6 & 35.0 & 18.4 & 44\tabularnewline
    \hfil 6 & \textbf{20.9} & \textbf{36.3} & \textbf{20.1} & \textbf{44.9}\tabularnewline
    \hline 
    \end{tabular}
    \caption{Model count ablation results}
    \label{tab:model_count_ablation}
\end{table}

\begin{table}[!ht]
    \centering
    \begin{tabular}{p{0.6cm}|p{0.8cm}p{0.8cm}p{0.8cm}p{0.8cm}}
    $k_{max}$ & $AP^{box}$ & $AP_{50}^{box}$ & $AP_{75}^{box}$ & $AR_{100}^{box}$\tabularnewline
    \hline 
    2 & 19.5 & 33.5 & 19 & 41.9\tabularnewline
    3 & \textbf{20.9} & \textbf{36.3} & \textbf{20.1} & 44.9\tabularnewline
    4 & 20.4 & 36.1 & 19.3 & \textbf{45.2}\tabularnewline
    5 & 19.4 & 35 & 18.1 & 44.9\tabularnewline
    \hline 
    \end{tabular}
    \caption{$k_{max}$ ablation results}
    \label{tab:k_max-ablation-results}
\end{table}

\begin{table}[!ht]
    \centering
    \begin{tabular}{p{0.6cm}|p{0.8cm}p{0.8cm}p{0.8cm}p{0.8cm}}
    $\tau^{m}$ & $AP^{box}$ & $AP_{50}^{box}$ & $AP_{75}^{box}$ & $AR_{100}^{box}$\tabularnewline
    \hline 
    0.1 & 17.7 & 30.6 & 17.1 & 37.2\tabularnewline
    0.2 & \textbf{20.9} & \textbf{36.3} & \textbf{20.1} & \textbf{44.9}\tabularnewline
    0.3 & 20.3 & 35.8 & 19.2 & 44.4\tabularnewline
    0.4 & 19.2 & 35.1 & 17.8 & 43.3\tabularnewline
    0.5 & 17.6 & 33.9 & 15.7 & 41.3\tabularnewline
    \hline 
    \end{tabular}
    \caption{$\tau^m$ ablation results}
    \label{tab:tau_m_ablation-results}
\end{table}

\section{Datasets}
\textbf{COCO and COCO20K} \cite{lin2014microsoft} is a large-scale instance segmentation and object detection dataset containing approximately $118K$ images for training and another $5K$ for validation. Additionally, COCO has an unannotated split of $123K$ images. 
COCO \texttt{20K} is a subset of the COCO \texttt{trainval2014} \cite{lin2014microsoft},
accommodated from 19817 randomly sampled images, used for evaluation in \cite{simeoni2021localizing, vo2020toward, wang2022tokencut, wang2023cut}. We evaluate our models in a class-agnostic manner on COCO \texttt{val2017} and COCO \texttt{20K}. We use COCO-style average precision (AP) and average recall (AR) from object detection and segmentation tasks for evaluation.
\\
\textbf{Pascal VOC} \cite{everingham2010pascal} 
is an object detection dataset widely used as a benchmark.  
\\
\textbf{OpenImages} V6 \cite{kuznetsova2020open}
Unifies instance segmentation, object detection and image classification, visual relationship detection, and more, in a single dataset. We evaluate our method on its $42K$ images from the \texttt{val} split. 
\\
\textbf{Clipart1k} \cite{inoue2018cross}
, which we refer to as 'Clipart', is an object detection dataset consisting of 1000 images from a clip art domain. We evaluate our model using all annotated images from this dataset, $traintest$.
\\
\textbf{Watercolor2K} \cite{inoue2018cross}
, which we refer to as 'Watercolor', is an object detection dataset consisting of 2000 images from a watercolor painting domain. We evaluate our model using all annotated images from this dataset, $traintest$.
\\
\textbf{Comic2K} \cite{inoue2018cross}
, which we refer to as 'Comic', is an object detection dataset consisting of 2000 images from a comic domain. We evaluate our model using all annotated images from this dataset, $traintest$.
\\
\textbf{LVIS} \cite{gupta2019lvis} (Large Vocabulary Instance Segmentation) is a large-scale dataset comprising $164K$ images featuring $2.2M$ high-quality instance segmentation masks. It covers over 1000 entry-level object categories, naturally forming a long-tail distribution of categories.

\section{Addtinal results}
\begin{table*}[b]
\centering
\begin{tabular}{p{2.4cm}|p{0.6cm}p{0.6cm}|p{0.6cm}p{0.6cm}|p{0.6cm}p{0.6cm}|p{0.6cm}p{0.6cm}|p{0.6cm}p{0.6cm}|p{0.6cm}p{0.6cm}|p{0.6cm}p{0.6cm}|}
\multirow{2}{*}{Method}&\multicolumn{2}{c|}{COCO}&\multicolumn{2}{c|}{COCO20K}&\multicolumn{2}{c|}{VOC}&\multicolumn{2}{c|}{OpenImages}&\multicolumn{2}{c|}{Clipart}&\multicolumn{2}{c|}{Watercolor}&\multicolumn{2}{c|}{Comic}\\ \cline{2-15}
&AP$_{75}$&AR&AP$_{75}$&AR&AP$_{75}$&AR&AP$_{75}$&AR&AP$_{75}$&AR&AP$_{75}$&AR&AP$_{75}$&AR\\ \hline
Prev. SOTA \cite{wang2023cut} &11.8&32.8&11.9&33.1&19.2&44&9.5&29.6&6&40.7&10.9&44.2&7.7&38.4\\
CuVLER (ours)&11.8&32.8&11.9&33&21.2&46.5&11.4&30.5&6.9&42&15.9&47.4&11.5&41\\
\textit{vs. prev. SOTA}&\textbf+0&\textbf+0&\textbf+0&\textbf-0.1&\textbf{+2}&\textbf{+2.5}&\textbf{+1.9}&\textbf{+0.9}&\textbf{+0.9}&\textbf{+1.3}&\textbf{+5}&\textbf{+3.2}&\textbf{+3.8}&\textbf{+2.6}\\
\hline
\end{tabular}
\caption{SOTA zero-shot unsupervised object detection performance on seven datasets. The reported results are based on the COCO metrics, encompassing both Average Recall (AR) and AP$_{75}$ scores. The presented models are trained in an unsupervised manner solely on ImageNet. Results of \cite{wang2023cut} are produced with official code and checkpoint. $AR$ refers to $AR^{box}_{100}$ metric. }
\label{tab:zeroshot-extra-results}
\end{table*}

On \Cref{tab:zeroshot-extra-results}, we can see additional results of a zero-shot evaluation of our CuVLER method and the results of CutLER, the previous SOTA. This table complement \Cref{tab:zeroshot}, with additional $AR^{box}_{100}$ and $AP^{box}_{75}$ metrics. We can see that our model suppressed CutLER across all the datasets in almost all metrics.


In \Cref{fig:supp_comparison}, we can visually observe the capabilities of VoteCut in generating superior detection and mask proposals, surpassing the previous SOTA methods. It is noteworthy that VoteCut succeeds in discovering multiple objects and assigns a confidence score to each proposal, a significant feature leveraged for subsequent training.

\begin{figure*}
    \centering
    \begin{subfigure}{0.3\linewidth}
        \caption*{TokenCut}
        \includegraphics[width=\linewidth]{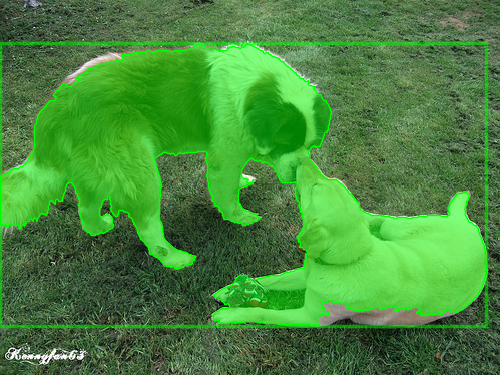}
        \\[1ex]
        \includegraphics[width=\linewidth]{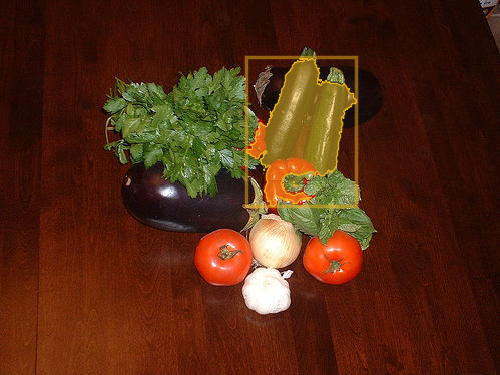}
        \\[1ex]
        \includegraphics[width=\linewidth]{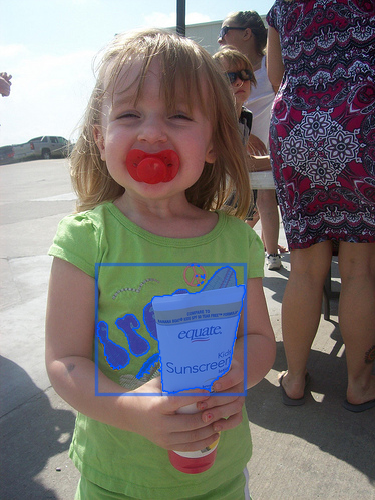}
        \\[1ex]
        \includegraphics[width=\linewidth]{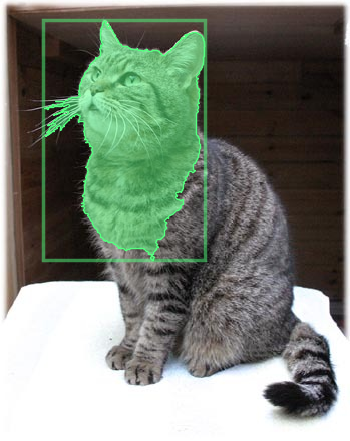}
    \end{subfigure}
    \hfill
    \begin{subfigure}{0.3\linewidth}
        \caption*{MaskCut}
        \includegraphics[width=\linewidth]{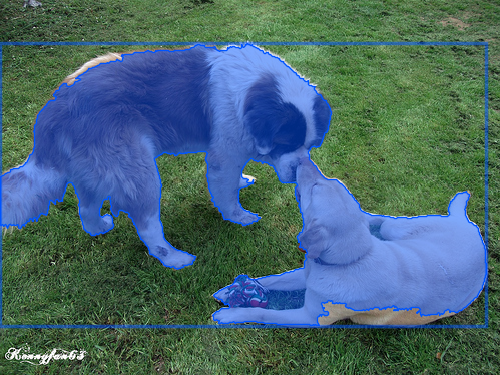}
        \\[1ex]
        \includegraphics[width=\linewidth]{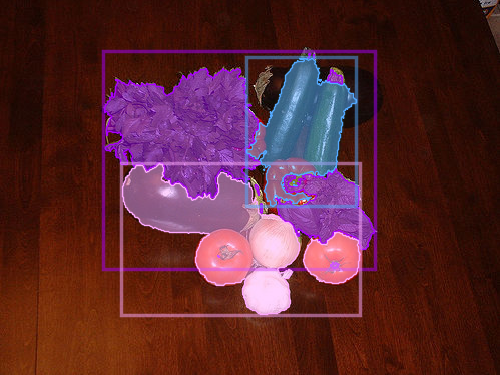}
        \\[1ex]
        \includegraphics[width=\linewidth]{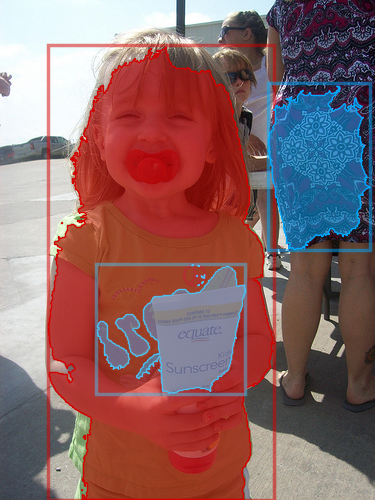}
        \\[1ex]
        \includegraphics[width=\linewidth]{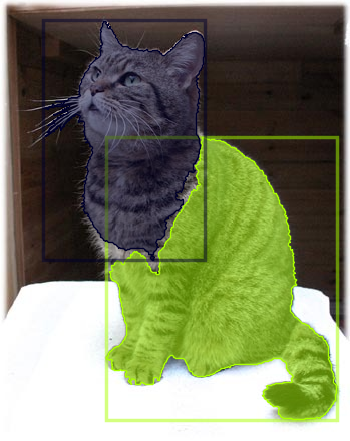}
    \end{subfigure}
    \hfill
    \begin{subfigure}{0.3\linewidth}
        \caption*{VoteCut (Ours)}
        \includegraphics[width=\linewidth]{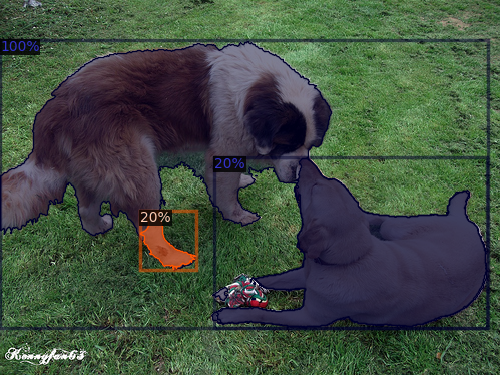}
        \\[1ex]
        \includegraphics[width=\linewidth]{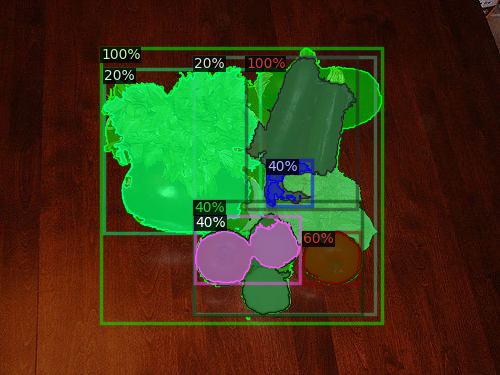}
        \\[1ex]
        \includegraphics[width=\linewidth]{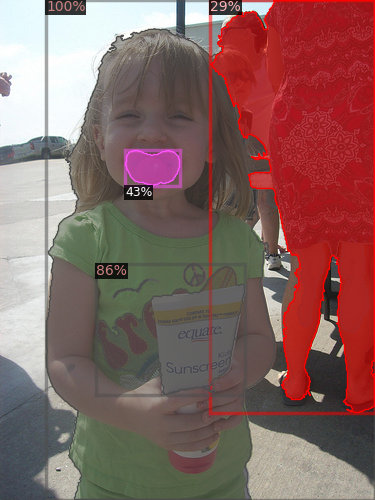}
        \\[1ex]
        \includegraphics[width=\linewidth]{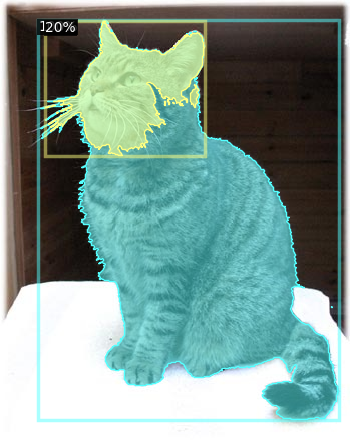}
    \end{subfigure}
\caption{Additinoal Visual illustration of VoteCut performance vs. SOTA NCut based object-discovery methods on the ImageNet validation set. The VoteCut bounding box score is calculated according to \cref{eq:score}}
\label{fig:supp_comparison}
\end{figure*}
\begin{figure*}
\ContinuedFloat
    \centering
    \begin{subfigure}{0.3\linewidth}
        \caption*{TokenCut}
        \includegraphics[width=\linewidth]{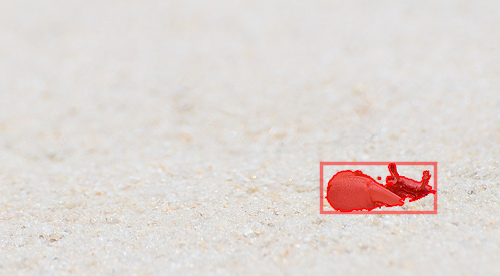}
        \\[1ex]
        \includegraphics[width=\linewidth]{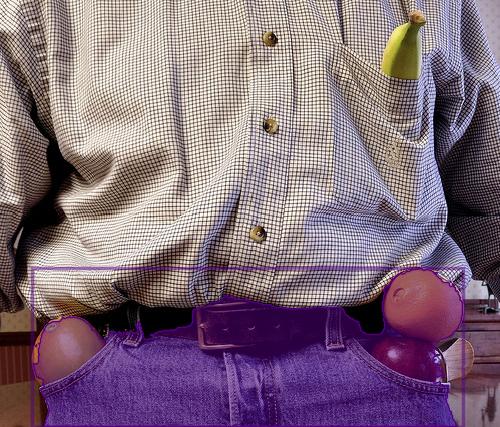}
        \\[1ex]
        \includegraphics[width=\linewidth]{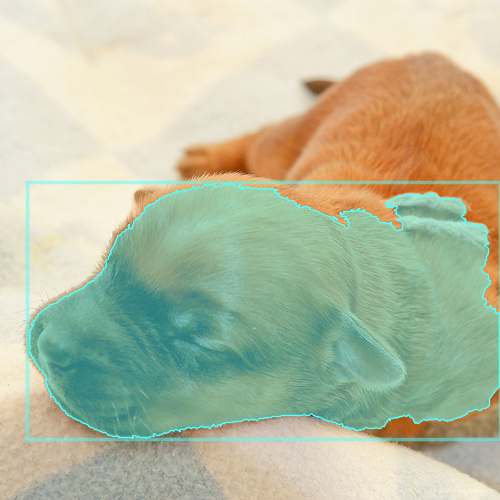}
        \\[1ex]
        \includegraphics[width=\linewidth]{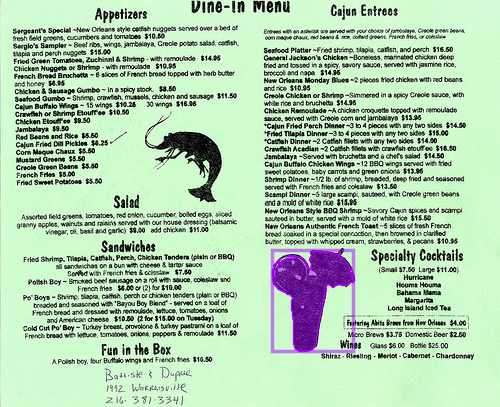}
        \\[1ex]
        \includegraphics[width=\linewidth]{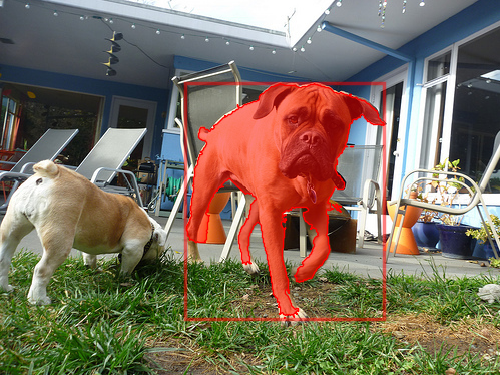}
    \end{subfigure}
    \hfill
    \begin{subfigure}{0.3\linewidth}
        \caption*{MaskCut}
        \includegraphics[width=\linewidth]{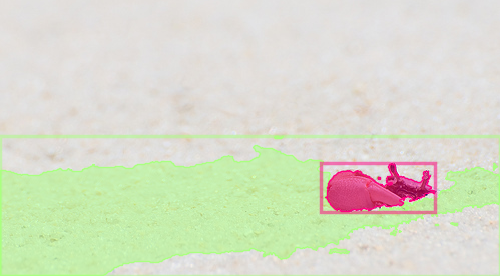}
        \\[1ex]
        \includegraphics[width=\linewidth]{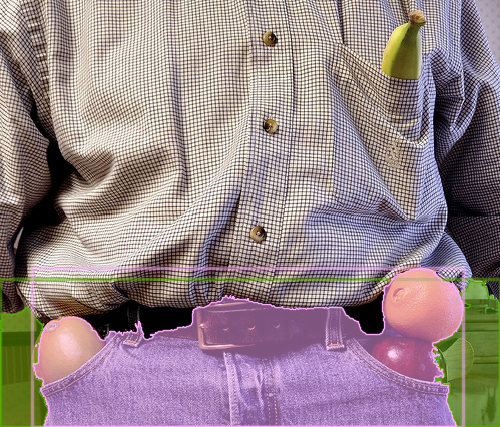}
        \\[1ex]
        \includegraphics[width=\linewidth]{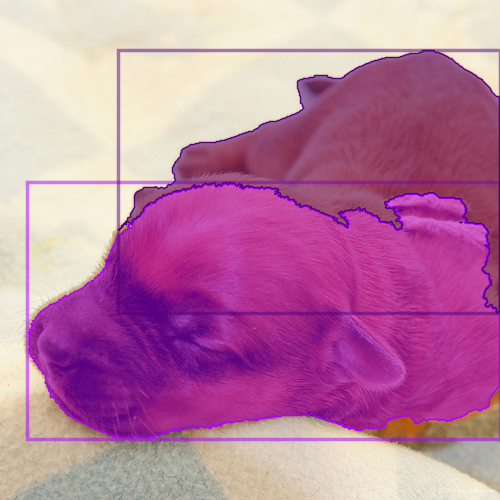}
        \\[1ex]
        \includegraphics[width=\linewidth]{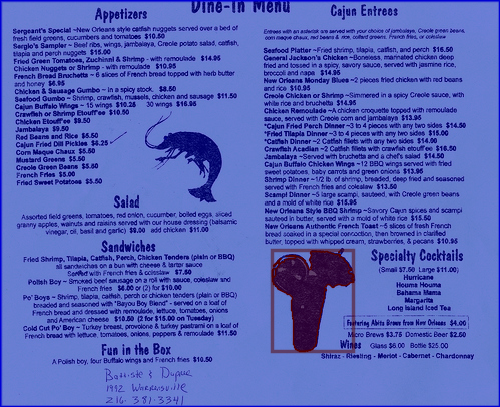}
        \\[1ex]
        \includegraphics[width=\linewidth]{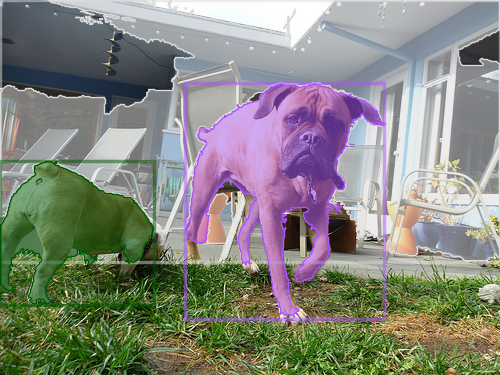}
    \end{subfigure}
    \hfill
    \begin{subfigure}{0.3\linewidth}
        \caption*{VoteCut (Ours)}
        \includegraphics[width=\linewidth]{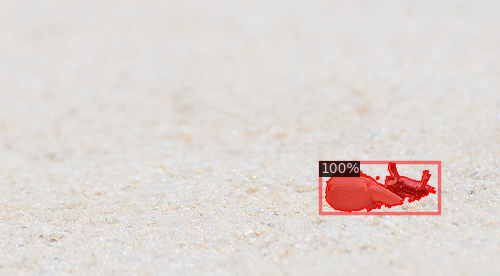}
        \\[1ex]
        \includegraphics[width=\linewidth]{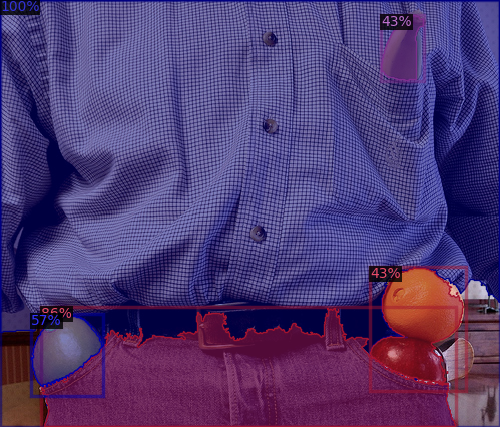}
        \\[1ex]
        \includegraphics[width=\linewidth]{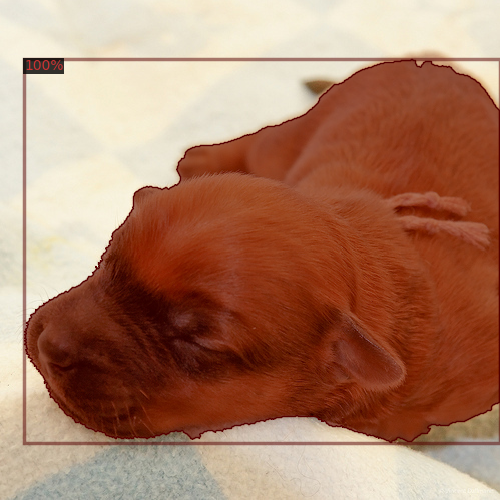}
        \\[1ex]
        \includegraphics[width=\linewidth]{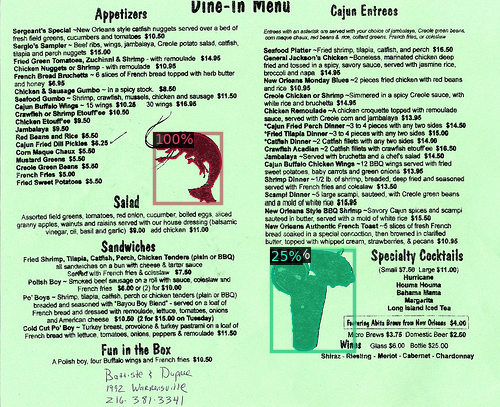}
        \\[1ex]
        \includegraphics[width=\linewidth]{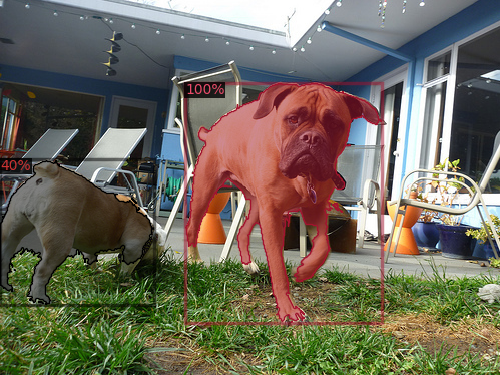}
    \end{subfigure}
\caption{Additinoal Visual illustration of VoteCut performance vs. SOTA NCut based object-discovery methods on the ImageNet validation set. The VoteCut bounding box score is calculated according to \cref{eq:score}}
\end{figure*}

\begin{figure*}
\ContinuedFloat
    \centering
    \begin{subfigure}{0.3\linewidth}
        \caption*{TokenCut}
        \includegraphics[width=\linewidth]{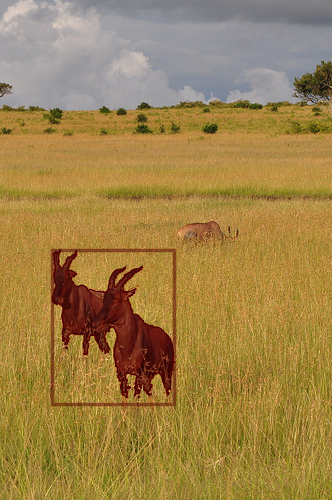}
        \\[1ex]
        \includegraphics[width=\linewidth]{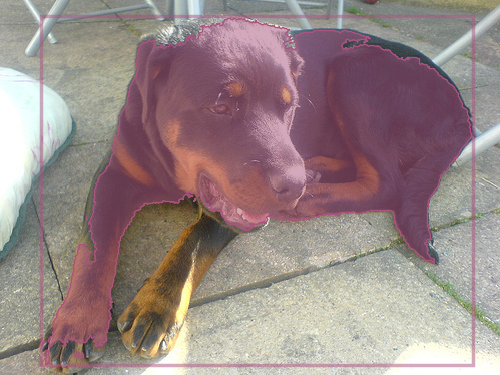}
        \\[1ex]
        \includegraphics[width=\linewidth]{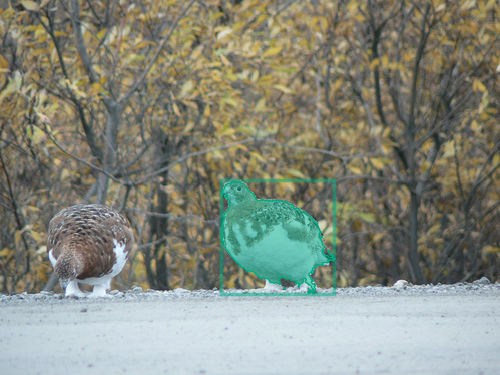}
    \end{subfigure}
    \hfill
    \begin{subfigure}{0.3\linewidth}
        \caption*{MaskCut}
        \includegraphics[width=\linewidth]{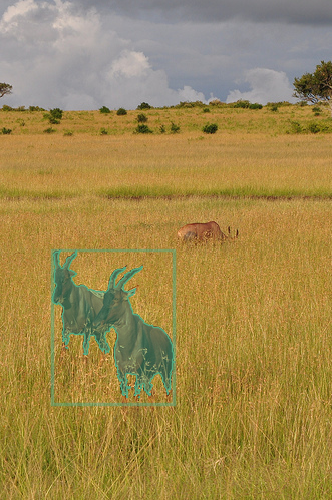}
        \\[1ex]
        \includegraphics[width=\linewidth]{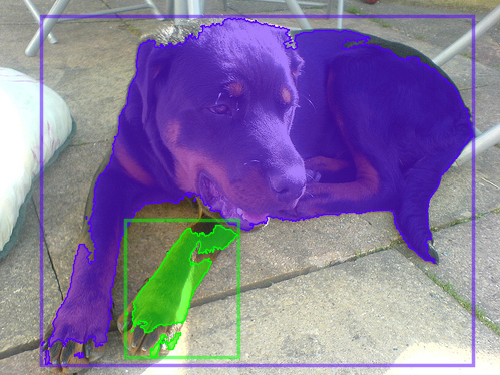}
        \\[1ex]
        \includegraphics[width=\linewidth]{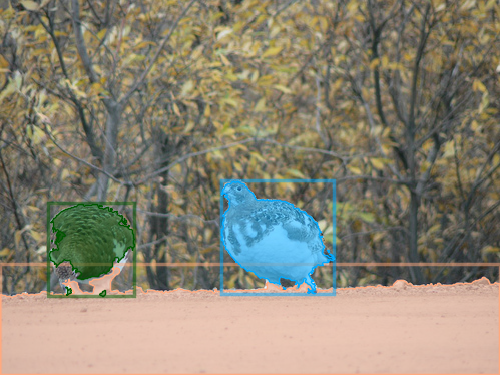}
    \end{subfigure}
    \hfill
    \begin{subfigure}{0.3\linewidth}
        \caption*{VoteCut (Ours)}
        \includegraphics[width=\linewidth]{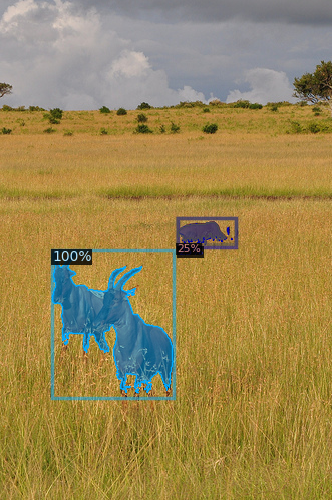}
        \\[1ex]
        \includegraphics[width=\linewidth]{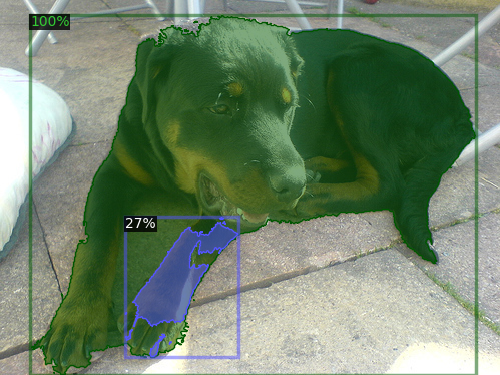}
        \\[1ex]
        \includegraphics[width=\linewidth]{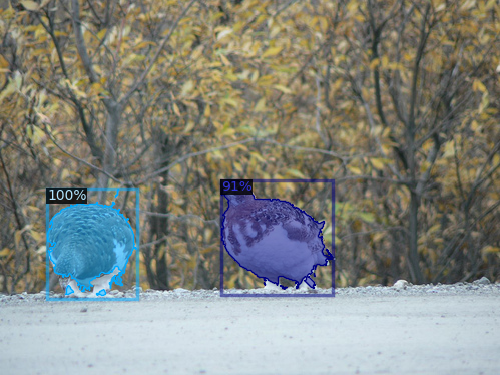}
    \end{subfigure}
\caption{Additinoal Visual illustration of VoteCut performance vs. SOTA NCut based object-discovery methods on the ImageNet validation set. The VoteCut bounding box score is calculated according to \cref{eq:score}}
\end{figure*}

\end{document}